\documentclass[10pt,a4paper,twoside]{article}

\input{spp2024.dat}

\begin{document}

\title{\TitleFont Classifying galaxies in the Galaxy10 DECals dataset using Inception and Residual CNNs}




 \author[1*]{Lanz Anthonee A.~Lagman}
 \author[2]{Prospero C. Naval, Jr.}
 \author[3]{Reinabelle C.~Reyes\lastauthorsep}
\affil[1]{Data Science Program, College of Science, University of the Philippines - Diliman, Philippines}
\affil[2]{Department of Computer Science, College of Engineering, University of the Philippines - Diliman, Philippines}
\affil[3]{National Institute of Physics, College of Science, University of the Philippines - Diliman, Philippines}
\affil[*]{\corremail{lalagman1@up.edu.ph} }

\begin{abstract}
\noindent
Image data regarding galactic morphology is expected to increase both in quantity and quality for the next foreseeable years; thus it is important to explore which deep learning architectures adapted for image classification tasks are cost-effective. 
Residual and Inception networks are ideal for exploring classification convolutional neural networks (CNNs) due to their computational efficiency, achieved through techniques such as residual connections and parallelized inception modules, enabling deeper networks without excessively increasing computational complexity.
In this work, we analyze the performance of ResNet101 and InceptionV4 on a spatially-augmented Galaxy10 DECals dataset. Retaining the ten-class classification of galaxies, we modify the image count of each class. 
We find that ResNet101 and InceptionV4 models achieved accuracies of $\sim90\%$, comparable with reported performance in the literature.
In terms of performance metrics, ResNet101 is superior to InceptionV4.
Our results indicate that either of these CNN architectures could serve as a robust foundation for specialized pipelines for classification of galaxy images from upcoming surveys.

\keywords{residual networks, galactic morphology, galaxy classification}

\end{abstract}

\maketitle
\thispagestyle{titlestyle}

\section{Introduction}\label{sec:intro}

The influx of data, both in quantity and quality is expected to exponentially increase yearly, as more advanced telescopes continue to start their operations and continuously gather data, which is often too much for astronomers to process. To give a brief background on the scale of data influx, the Hubble Space Telescope (HST), which was launched in 1990, sends one to two gigabytes of data daily. The James Webb Space Telescope (JWST), HST's successor, which was launched in 2021, is currently sending 50 gigabytes daily \cite{futurismAstronomersComplain}.

\begin{figure}[!ht]
  \centerline{\includegraphics[width=1\linewidth]{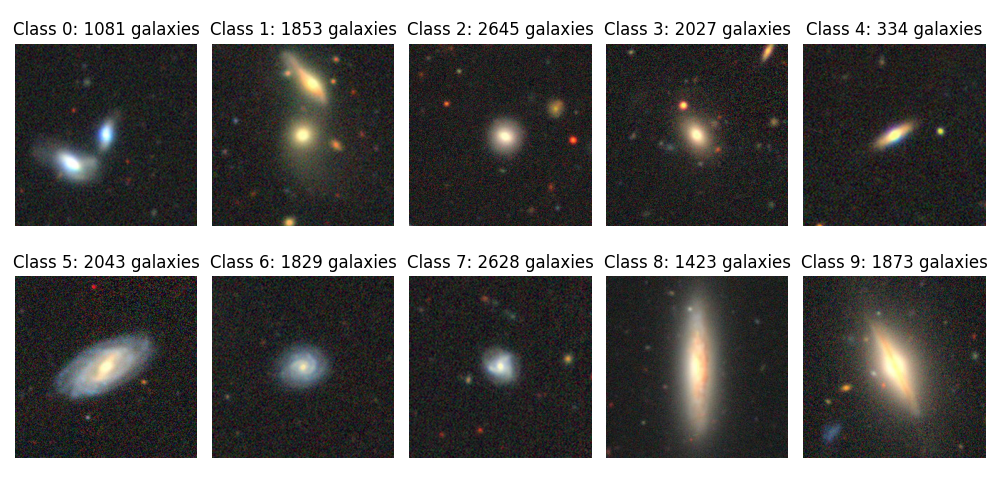}}
  \caption{Galaxy10 DECals dataset classes and respective number of images per class.}
  \label{fig:loss_curves} 
\end{figure}

\noindent While there are several notable image datasets derived from space telescopes across the past decades, we need a large, labeled dataset for this study. 
We use the Galaxy10 DECals dataset \cite{Galaxy10DECals}, which is composed of 17,736 RGB images of size $256 \times 256$, with the colors representing the r, g, and z-bands in the optical, respectively. 
The images were sourced from DESI Legacy Image Survey \cite{DESI}, which is the combination of projects from the Dark Energy Camera Legacy Survey, the Beijing-Arizona Sky Survey, and the Mayall z-band Legacy Survey. 
Meanwhile, the labels were sourced from the Galaxy Zoo project \cite{GalaxyZoo}. 

As with CIFAR10 \cite{CIFAR10}, galaxy image datasets such as Galaxy10 DECals are suitable testing grounds for comparing and modifying different convolutional neural network (CNN) architectures. In the case of other well-known CNN architectures, four variations of ResNet have been used by Alawi et. al (2021) \cite{Classify_ResNet} to perform binary classification on images of stars and galaxies with high accuracies of around 90\%. Chen (2023) \cite{Classify_VGG} used VGG16 to classify all 17,736 from Galaxy10 DECals dataset while retaining the ten-class classification scheme and achieved an accuracy of 81\%. The Galaxy10 DECals dataset has also been used to train several variations of pretrained DenseNet by Wuyu et al (2022).
\cite{Classify_DenseNet} They found that DenseNet121 outperforms the other models and achieved an accuracy of 89\%. 
In this work, we will assess the performance of popularly-used CNNs for classification, ResNet101 and InceptionV4, on the Galaxy10 DECals dataset.

\section{Methodology}\label{sec:methodology}


\subsection{Residual and Inception Networks}

ResNet101 is the 101-layer version of ResNet, a type of CNN architecture that introduces the deep residual learning framework. Introduced by He et al. \cite{ResNet}, the network uses residual blocks in order to solve the vanishing gradient problem, which exposes degradation problems. A degradation problem happens when the depth of a network increases; its accuracy becomes saturated, and therefore, degrades rapidly. 
Composed of convolutional and batch normalization layers and ReLu activation function, residual blocks solve these by including shortcut connections that bypass one or more convolutional layers. These shortcuts enable the network to learn residual functions, capturing the difference between input and output. 

InceptionV4, presented by Szegedy et al. \cite{InceptionV4}, serves as the improved but simplified version of InceptionV3 by simplifying and making modules more uniform. It achieved this by modifying the stem and introducing specialized Reduction Blocks to explicitly alter the width and height of the feature grid, improving feature extraction efficiency compared to earlier versions. 
It still uses its Inception module, which is a parallelization of convolution modules of different filter sizes that capture features across different granularities.

\subsection{CNN Training Setup}

For the hyperparameters, all models were identically trained with batch size of 64, 150 epochs, learning rate of 0.001, CrossEntropy Loss function, and Adam optimizer. As for hardware, the model optimization and testing were conducted on an NVIDIA RTX A4000 GPU. 
To minimize class bias, we performed both offline and online augmentation. 
Offline augmentation is performed on the dataset itself before the neural networks are trained, while online augmentation is performed on-the-fly, i.e. they are dynamically applied during training.

\subsection{Data Transformation and Augmentation}

In the offline augmentation, we set each image class population to be equal to 2,500; adding images via data augmentation to all classes except for class 2, for which we retained only the first 2,500 images. 
The spatial transformations that we performed were random orthogonal rotations and flips that are either vertical or horizontal. 

In the online augmentation for data augmentation, the spatial transformations are the same as with the offline data augmentation for the train set. 
In addition, the images are resized to $224\times224$ pixels, normalized using global mean and standard deviation values, and converted to PyTorch tensors. 
For the validation and test sets, these same transformations are applied, but no spatial transformations are performed. 

\section{Results and Discussion}\label{sec:results}

\subsection{Training Results}

The loss curves for ResNet101 and InceptionV4 models are shown in Figure \ref{fig:loss_curves} (left and right panels, respectively). 
The validation loss curve has already turned over within 150 epochs and the minimum loss corresponds to epoch 46 for ResNet101 and epoch 71 for InceptionV4.
Table \ref{tab:training_results} compares the training runtime, training and validation accuracies of the two CNNs.
ResNet101 has longer training runtime but it achieved higher accuracies.

\begin{figure}[!ht]
  \centerline{\includegraphics[width=1\linewidth]{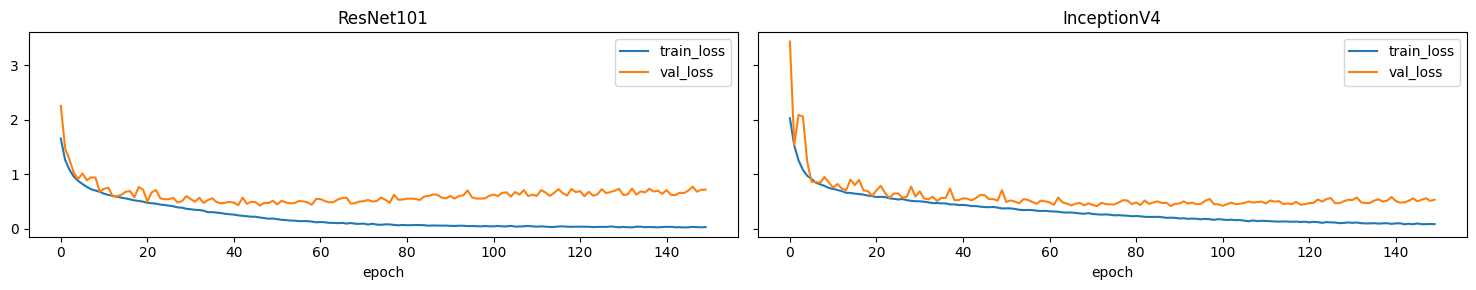}}
  \caption{Loss curves from training and validation of the selected CNN architectures.}
  \label{fig:loss_curves} 
\end{figure}


\begin{table}[!ht]
\centering
\begin{tabular}{|l|c|c|c|}
\hline
\textbf{CNN} & \textbf{Runtime (H:M:S)} & \textbf{Training Accuracy} & \textbf{Validation Accuracy}\\
\hline
ResNet101 & 04:41:57 & $92.91 \pm 9.33\%$ & $84.85 \pm 7.94\%$ \\
InceptionV4 & 03:59:51 & $88.21 \pm 10.15\%$ & $82.40 \pm 9.93\%$ \\
\hline
\end{tabular}
\caption{Training runtime and accuracy of the CNN models on Galaxy10 DECals dataset across 150 epochs.}
\label{tab:training_results}
\end{table}
\subsection{Test Results}

Now, we will compare how well the models performed on the test set.
Table \ref{tab:test_metrics} lists the test performance metrics of the two CNNs.
We find that both models achieved accuracies of $\sim90\%$, comparable or greater than other models in the literature.
Comparing the two, ResNet101 achieved slightly better performance across all metrics, although this difference may not be statistically significant.

\begin{figure*}[!ht] 
    \centering
  \centerline{\includegraphics[width=1\linewidth]{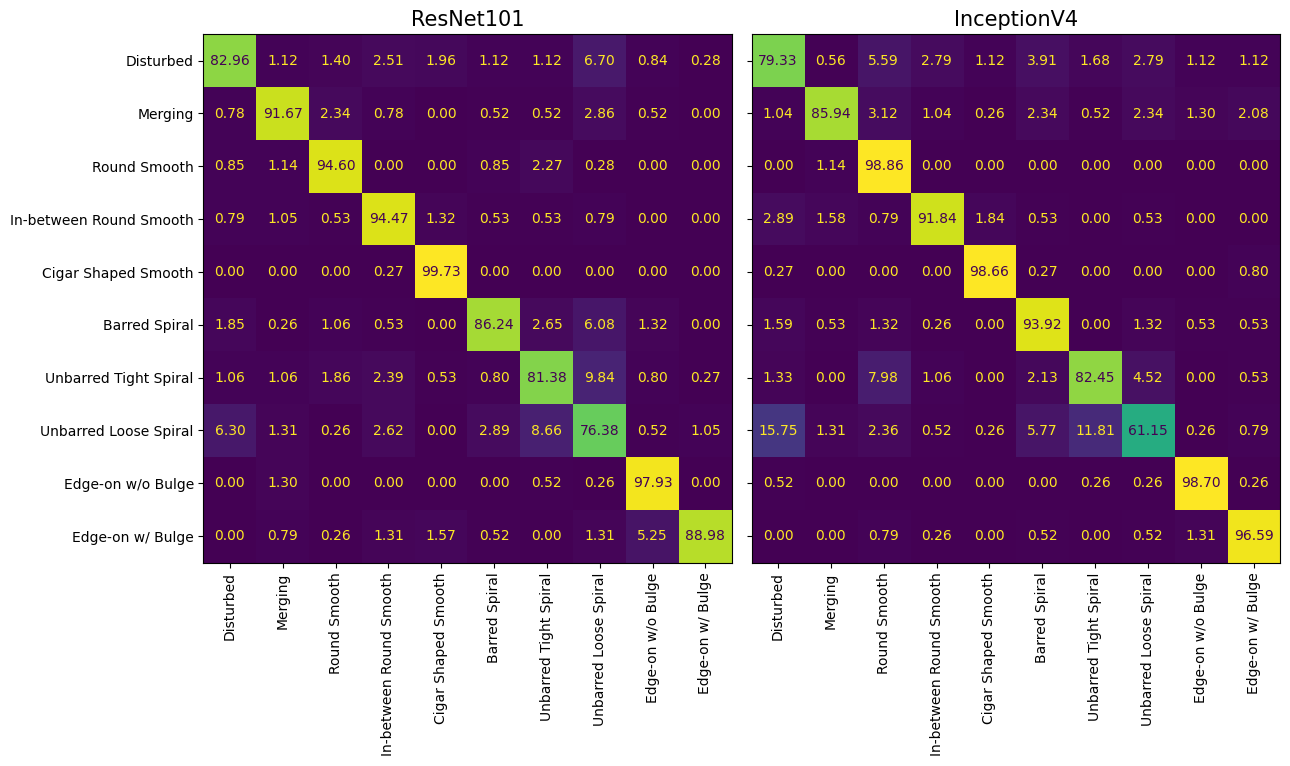}}
  \caption{Normalized confusion matrices for ResNet101 and InceptionV4, showing accuracies per class in percentages. The x-axis indicates the predicted labels while the y-axis indicates the true labels.}
  \label{fig:confusion_matrices}
\end{figure*}

The confusion matrices for both models are shown in Figure \ref{fig:confusion_matrices}. For ResNet101, the largest misclassifications are between unbarred tight spirals and unbarred loose spiral galaxies; while for InceptionV4, it is between unbarred loose spirals and disturbed galaxies. ResNet101 was able to classify cigar-shaped smooth galaxies with 99.73\% accuracy, and also did very well with edge-on galaxies without bulge with 97.93\% accuracy. 
InceptionV4, meanwhile, was able to classify round smooth, barred spiral, unbarred tight spiral galaxies, and edge-on galaxies with and without bulge better than ResNet101. 

\begin{table}[!ht]
\centering
\begin{tabular}{|l|c|c|c|c|}
\hline
\textbf{CNN} & \textbf{Accuracy} & \textbf{Precision} & \textbf{Recall} & \textbf{F1} \\
    \hline
    ResNet101 & 0.8944 & 0.895356 & 0.894382 & 0.894197 \\
    InceptionV4 & 0.8872 & 0.886657 & 0.887499 & 0.883960 \\
    \hline
\end{tabular}
\caption{Performance metrics of the CNN models on the Galaxy10 DECals dataset.}
\label{tab:test_metrics}
\end{table}


\section{Conclusions and Recommendation}
We have implemented CNN architectures ResNet101 and InceptionV4, for a multi-class classification task on a spatially-augmented galaxy image dataset in order to assess how well they perform.
Overall, we find that ResNet101 is the superior CNN between the two, achieving an overall accuracy of 89.5\% and accuracies exceeding 75\% across all categories. In addition, while it required a longer training time for the same number of epochs (by around 20\%), it achieved its minimum validation loss earlier; in practice, it would take a shorter training time (86 vs. 113 minutes) if early stopping is implemented. 

While we recognize that there are limitations to this study-- there can be potential biases in the dataset (which is based on crowd-sourced labels), and we have not included a wider variety of CNN architectures (such as VGG, EfficientNet, DenseNet, and combinations of inception and residual networks, such as V1 or V2 of Inception-ResNets), the results of this experiment serve as a stepping stone for future work. 
Test performance could be evaluated several times in order to assess robustness and further error analysis could also be performed by investigating misclassified galaxies and their properties.
\bibliographystyle{spp-bst}
\bibliography{bibfile}

\end{document}